\pdfoutput=1
\documentclass[11pt]{article}

\usepackage[preprint]{acl}
\usepackage{amssymb}  
\usepackage{array}    % 引入array宏包，用于调整线条粗细
\DeclareUnicodeCharacter{0259}{e}
\usepackage{times}
\usepackage{enumitem}  
\usepackage{latexsym}
\usepackage{booktabs}
\usepackage{multirow}
\usepackage{amsmath}
\PassOptionsToPackage{table,xcdraw}{xcolor}
\usepackage{xcolor}
\usepackage[T1]{fontenc}
\usepackage{microtype}
\usepackage{graphicx}
\usepackage{subfigure}
\usepackage{threeparttable}
\usepackage{multirow}
\usepackage{booktabs} % for professional tables
\usepackage{stfloats}
\usepackage[utf8]{inputenc}
\usepackage{microtype}
\usepackage{threeparttable}
\usepackage{inconsolata}

\usepackage{graphicx}
\usepackage{algorithm}
\usepackage{algpseudocode}
\usepackage{setspace}

\title{DR-RAG: Applying Dynamic Document Relevance to Retrieval- \\ Augmented Generation for Question-Answering
}
\author{Zijian Hei$^{*,1,2}$, Weiling Liu$^{*,1,3}$, Wenjie Ou$^{*,1,4}$, Juyi Qiao$^{1}$, Junming Jiao$^{1}$, \\ \bf{Guowen Song$^{\footnotemark[2],1}$, Ting Tian$^{\footnotemark[3], 2}$, Yi Lin$^{\footnotemark[3], 4}$} \\
  $^1$ Li Auto Inc.,$^2$ Sun Yat-sen University,$^3$ Northeastern University, China,$^4$ Sichuan University \\
  \texttt{songguowen@lixiang.com} \\
  \texttt{tiant55@mail.sysu.edu.cn}, \texttt{yilin@scu.edu.cn}
 }

\begin{document}
\maketitle
\begin{abstract}
\renewcommand{\thefootnote}{\fnsymbol{footnote}}
\footnotetext[1]{Contribute equally during internshiping at Li Auto.}
\footnotetext[2]{Corresponding author 1.}
\footnotetext[3]{Corresponding author 2.}
Retrieval-Augmented Generation (RAG) has recently demonstrated the performance of Large Language Models (LLMs) in the knowledge-intensive tasks such as Question-Answering (QA). RAG expands the query context by incorporating external knowledge bases to enhance the response accuracy. However, it would be inefficient to access LLMs multiple times for each query and unreliable to retrieve all the relevant documents by a single query. We have found that even though there is low relevance between some critical documents and query, it is possible to retrieve the remaining documents by combining parts of the documents with the query. To mine the relevance, a two-stage retrieval framework called \textbf{D}ynamic-\textbf{R}elevant \textbf{R}etrieval-\textbf{A}ugmented \textbf{G}eneration (\textbf{DR-RAG}) is proposed to improve document retrieval recall and the accuracy of answers while maintaining efficiency. Additionally, a compact classifier is applied to two different selection strategies to determine the contribution of the retrieved documents to answering the query and retrieve the relatively relevant documents. Meanwhile, DR-RAG call the LLMs only once, which significantly improves the efficiency of the experiment. The experimental results on multi-hop QA datasets show that DR-RAG can significantly improve the accuracy of the answers and achieve new progress in QA systems.
\end{abstract}

\section{Introduction}

Large language models (LLMs) have recently made significant improvement in the field of Natural Language Processing (NLP), especially in text generation tasks \cite{brown2020language, achiam2023gpt, touvron2023llama2, anil2023palm, ouyang2022training, touvron2023llama}. Although LLMs excel in various application scenarios, challenges remain regarding the accuracy and timeliness of the generated text, especially in real-time domains. LLMs with intrinsic parameter memories may generate inaccurate or even incorrect text when faced with up-to-date query \cite{min2023factscore, mallen2022not, muhlgay2023generating}. This issue, known as hallucination, occurs when the text generated by LLMs fails to align with real-world knowledge \cite{ji2023surveyhallucination, zhang2023sirenhallucination, kwiatkowski2019naturalquestion}. Therefore, Retrieval-Augmented Generation (RAG) frameworks have been proposed to improve the accuracy of generated text by combining relevant information from external knowledge base with query \cite{arora2023gar, lewis2020retrieval, borgeaud2022improving}. RAG has effectively demonstrated its superiority in knowledge-intensive tasks such as open-domain Question-Answering (QA) and has achieved new progress in the LLMs' performance.

\begin{figure}[t]
    \centering
    \includegraphics[width=1\linewidth]{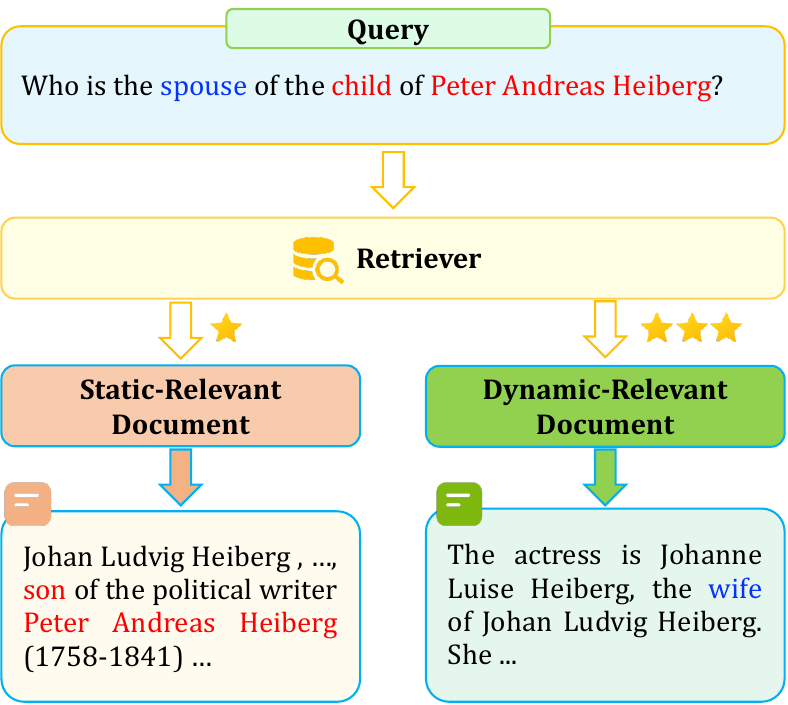}
        % \caption{Caption 1}
    \caption{An example shows that retriever easily introduces static-relevant documents due to high relevance (\textcolor{red}{red}), but struggles to retrieve dynamic-relevant documents which are of low relevance (\textcolor{blue}{blue}) but critical for the answer. Stars are levels of retrieval difficulty.}
    \label{fig:def}
    \vspace{-0.4cm}
\end{figure}

However, irrelevant information reduces the quality of the generated text and further interferes with the ability of LLMs to answer the query in the application \cite{shi2023large}. Moreover, the undifferentiated combining strategy in RAG can lead to mixing in some irrelevant information \cite{rony2022dialokg}. Inconsistent or contradictory information during combining the document may lead to the introduction of incorrect information and have an impact on the accuracy of the generated answers. In the retrieval, we need to select documents that are highly relevant and decisive for the generation of answers (\textit{static-relevant documents}) and documents that are low relevant but also crucial to the generation of answers (\textit{dynamic-relevant documents}). As shown in Fig. \ref{fig:def}, an example query is ‘Who is the spouse of the child of Peter Andreas Heiberg?’, which requires the two most relevant documents to obtain the correct answers. Static-relevant documents is easy to be retrieved due to the high relevance with the query on ‘Peter Andreas Heiberg’ and ‘child/son’ (Fig. \ref{fig:def} \textcolor{red}{red}). However, dynamic-relevant documents is difficult to be retrieved because it is only related to the query as a ‘spouse/wife’ (Fig. \ref{fig:def} \textcolor{blue}{blue}). Moreover, the knowledge base contains too much information about ‘spouse’, which may cause dynamic-relevant documents to be ranked lower in the retrieval process. There is a high relevance on ‘Johan Ludvig Heiberg’ and ‘wife’ between static- and dynamic-relevant documents. If ‘spouse/wife’ with the query is also taken into account, we can easily retrieve dynamic-relevant documents to get the answer.

Motivated by the above observations, a novel two-stage retrieval framework called \textbf{D}ynamic-\textbf{R}elevant \textbf{R}etrieval-\textbf{A}ugmented \textbf{G}eneration (\textbf{DR-RAG}) is proposed to mine the relevance between the query and documents. In the first-retrieval stage, similarity matching (SM) method is used to obtain a certain percentage of documents based on the query. Subsequently, the documents with the query are concatenated to dig further into more in-depth relevance to dynamic-relevant documents. Moreover, we design a classifier that determines whether the retrieved documents contribute to the current query by a predefined threshold. To optimise the documents, we design two approaches, i.e., forward selection and reverse selection. We aim to ensure that the retrieved documents are highly relevant, thus avoiding redundant documents. Through two-stage retrieval and classifier selection strategies, DR-RAG has the ability to retrieve sufficient relevant documents and address complex and multilevel problems. DR-RAG can make full use of the static and dynamic relevance of documents and enhance the model's performance under diverse queries. To validate the effectiveness of DR-RAG, we conduct extensive experiments by different retrieval strategies on multi-hop QA datasets. The results show that our method can significantly improve the recall and accuracy of the answers.

In short, we summarize the key contributions of this work as follows:
\begin{itemize}
\item We design an effective RAG framework named DR-RAG, which is effective in multi-hop QA. Two-stage retrieval strategy is proposed to significantly improve the recall and accuracy of the retrieval results.
\item We design a classifier that determines whether the retrieved documents contribute to the current query by setting a predefined threshold. The mechanism can effectively reduces redundant documents and ensures that the retrieved documents are concise and efficient.
\item We conduct experiments on three multi-hop QA datasets to validate our DR-RAG. The experimental results show that DR-RAG has the ability to improve recall by 86.75\%  and improve by 6.17\%, 7.34\%, 9.36\% in the three metrics (Acc, EM, F1). DR-RAG has significant advantages in complex and multi-hop QA and support the performance of the RAG frameworks in QA systems.
\end{itemize}

\begin{table*}[t]
\footnotesize
\renewcommand{\arraystretch}{1.1}%调行距
\centering
\setlength{\tabcolsep}{1pt}
\linespread{1}
\caption{The key mathematical notations.}
\label{tab:my-table math}
\begin{tabular}{@{}cl@{}}
\toprule
Notation                                         & \multicolumn{1}{c}{Description}                                                 \\ \midrule
$\boldsymbol{q}$                                            & the user’s input query             \\   
$\boldsymbol{q}^*$                                            & 
the query combined with retrieved document after the first-retrieval stage
                                                                 \\
$D$                                            & the knowledge base for storing documents                                         \\
$C$                                            & the trained classifier                                                      \\
$k$                                            & the total number of documents retrieved from $D$                            \\
$k_1$                                         & the total number of documents retrieved from $D$ in the first-retrieval stage  \\
$k_2$                                         & the total number of documents retrieved from $D$ in the second-retrieval stage \\
$n$                                            & the number of documents critical for correctly answering $\boldsymbol{q}$                   \\
$\boldsymbol{d}$                                            & the documents retrieved from $D$                                              \\
$\boldsymbol{d}^*$                         & the relevant documents for correctly answering $\boldsymbol{q}$                           \\
$\boldsymbol{d}^\Delta$ & the irrelevant documents for correctly answering $\boldsymbol{q}$                         \\
$\boldsymbol{d_{stat}^*}$               & the documents with static relevance to the query    \\
$\boldsymbol{d_{dyn}^*}$               & the documents with dynamic relevance to the query \\ \bottomrule
\end{tabular}
\vspace{-0.4cm}
\end{table*}

\section{Method}

In this section, we will describe the DR-RAG framework and its design approach in detail. Specifically, in section \ref{3.1} we will define relevant symbols comprehensively, and in section \ref{3.2} we will describe the whole framework.

\begin{figure*}[t]
    \centering
    \includegraphics[width=0.8\linewidth]{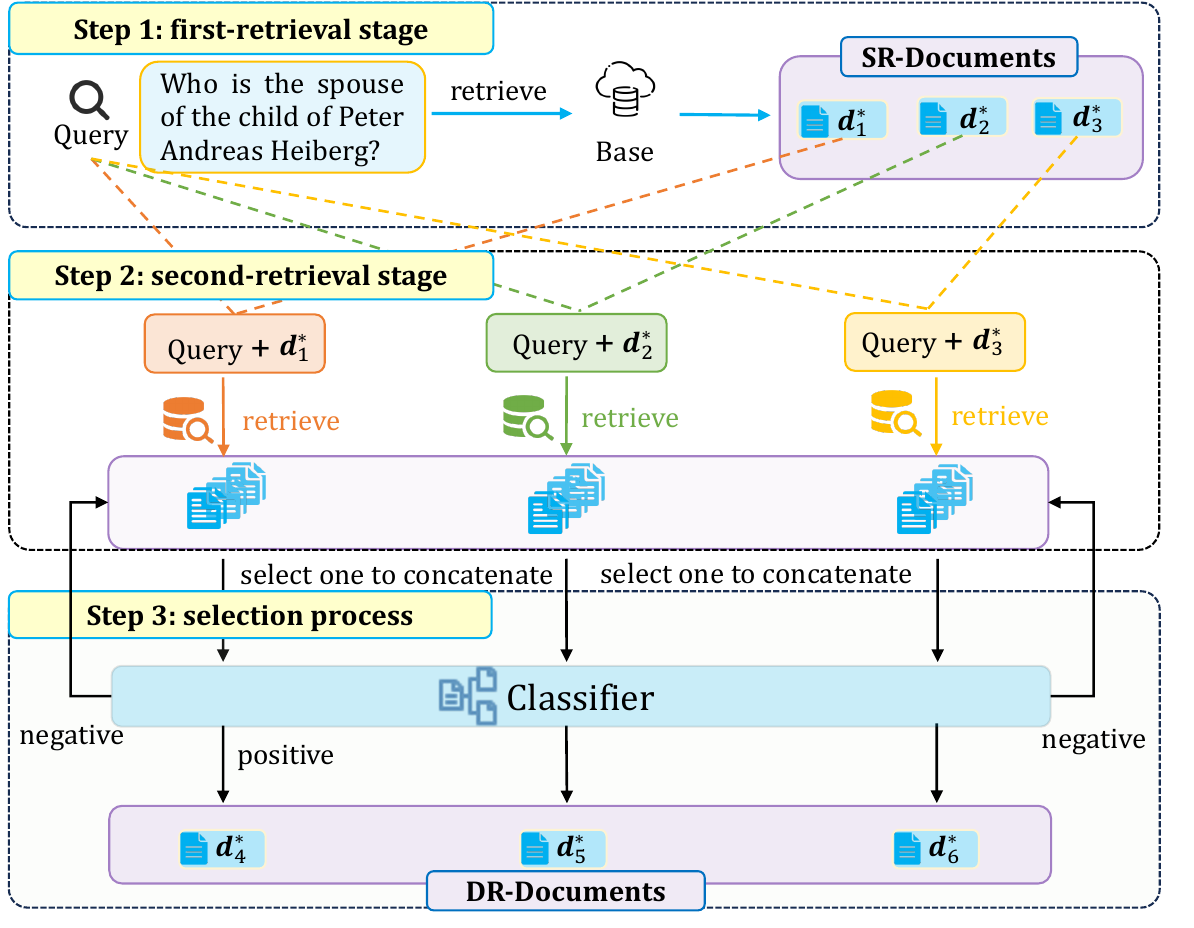}
        % \caption{Caption 1}
    \caption{An overview of DR-RAG. In step 1, we retrieve static-relevant documents (SR-Documents) due to high relevance with the query. Then we concatenate SR-Documents with the query to retrieve multiple dynamic-relevant documents (DR-Documents) in step 2. Finally, we select each of DR-Documents in turn to concatenate with the query and SR-Documents and feed them into the classifier to select the most relevant DR-Document.}
    \label{fig:model}
    \vspace{-0.4cm}
\end{figure*}

\subsection{Preliminaries}
\label{3.1}
To enrich the knowledge of LLMs, we need to retrieve multiple documents to provide comprehensive answers to complex query. For better clarity, we summarize the key notations in Table \ref{tab:my-table math} and the whole framework can be referred to in Fig. \ref{fig:model}.

Our goal is to retrieve the most relevant documents $\boldsymbol{d}^*$ from the retrieved documents $\boldsymbol{d}$ to answer the query and prevent missing key information from the additional knowledge provided to LLMs. However, it is difficult to retrieve all the static- and dynamic-relevant documents through SM method during the retrieval process (Fig. \ref{fig:model}). For clearness, we name these two types of relevant documents as $\boldsymbol{d_{stat}^*}$ and $\boldsymbol{d_{dyn}^*}$, respectively. 

A common approach is to increase the value of $k$ to expand the possibility of retrieving $\boldsymbol{d_{dyn}^*}$. For instance, in MuSiQue, increasing $k$ from 3 to 6 only raises the recall rate from 58\% to 76\%, leaving many relevant documents unretrieved. Furthermore, irrelevant documents will provide LLMs with redundant information. Motivated by the problem, the main research objective of our work is to improve the document recall rate of $\boldsymbol{d_{dyn}^*}$ based on dynamic relevance with the same top-$k$.

\subsection{DR-RAG}
\label{3.2}
In this section, we will give a comprehensive description about the DR-RAG framework, a new two-stage retrieval method compared to traditional reranking methods \cite{youdao_bcembedding_2023, chen2024bge}. From Fig. \ref{fig:model}, we retrieve $k_1$ documents through SM method (first-retrieval stage) and employ a classifier $C$ to model the dynamic relevance between documents (selection process) to enhance the recall rate of the remaining $k_2$ documents. The classifier $C$ lies in assessing the dynamic relevance between documents to determine whether the information from the documents is crucial to answer the present query.

\subsubsection{Query Documents Concatenation}
As mentioned before, due to the low relevancy between dynamic-relevant documents and the query, the documents are difficult to be retrieved. Moreover, the only relevant information ‘spouse/wife’ between them is also obscured by the mixed information in the knowledge base because too many documents in $D$ will contain ‘spouse’. Therefore, Query Documents Concatenation (QDC) method aims to employ the sentence to match for more useful and relevant information. After the first-retrieval stage, we will obtain $k_1$ static-relevant documents and concatenate $\boldsymbol{q}$ with each document to form multiple <$\boldsymbol{q}, \boldsymbol{d_i}, i \in k_1$> pairs. Moreover, dynamic-relevant documents from $D$ can be retrieved by corresponding <$\boldsymbol{q}, \boldsymbol{d_i}, i \in k_1$> pair in the second-retrieval stage. As the case in Fig \ref{fig:model}, when $\boldsymbol{q}$ and $\boldsymbol{d_{stat}^*}$ are concatenated, the query contains both the ‘Johan Ludvig Heiberg’ and the relationship ‘spouse/wife’, which is essentially similar to $\boldsymbol{d_{dyn}^*}$. Therefore, $\boldsymbol{d_{dyn}^*}$ is more clearly related to the query and thus easily retrieved. The whole process is:
\begin{align}\footnotesize
\label{qdc}
\begin{split}
    Cnt &= \{\} \\
    \{\boldsymbol{d}_1, \boldsymbol{d}_2, \ldots, \boldsymbol{d}_{k_1}\} &= \texttt{Retriever}(\boldsymbol{q}) \\
    Cnt &= Cnt \cup \{\boldsymbol{d}_1, \boldsymbol{d}_2, \ldots, \boldsymbol{d}_{k_1}\} \\
    \boldsymbol{q}_i^* &= \texttt{Concat}(\boldsymbol{q}, \boldsymbol{d}_i) \\
    \{\boldsymbol{d}'_{i,1}, \ldots, \boldsymbol{d}'_{i,k_2}\} &= \texttt{Retriever}(\boldsymbol{q}_i^*) \\
    Cnt &= Cnt \cup \{\boldsymbol{d}'_{i,j} \mid \boldsymbol{d}'_{i,j} \not\in Cnt \land first\} \\
    answer &= \texttt{LLM}(\texttt{Concat}(\boldsymbol{q}, Cnt))
\end{split}
\end{align}
where $k_1+k_2$ is equal to $k$. $\texttt{Retriever}$ is a common SM method. $\boldsymbol{d}$ and $\boldsymbol{d'}$ are the relevant document retrieved from $D$ in the first and second-retrieval stage. $Cnt$ is a context containing multiple documents. $Cnt = Cnt \cup \{\boldsymbol{d}'_{i,j} \mid \boldsymbol{d}'_{i,j} \not\in Cnt \land first\}$ means that for for a given $\boldsymbol{d'}$, the first $\boldsymbol{d}'_{i,j}$ in the second-retrieval stage that is not already part of $Cnt$ will be placed into $Cnt$. $\texttt{LLM}$ is a large language model to obtain the answer. $answer$ is the output to answer the query.

\subsubsection{Classifier for Selection}
While QDC method significantly improves document recall and answer accuracy, there are two key issues to consider: 1) There may be redundant information in the $k$ retrieved documents, which may affect the response of LLMs; 2) How to determine whether a document retrieved in the second-retrieval stage is valid for an answer to further optimise document recall. Motivated by the issues, two pipelines are designed to dig into in-depth document relevance and solve the issues: 1) Classifier Inverse Selection (CIS): in this pipeline, after the second-retrieval stage we exclude some irrelevant documents from the $k$ retrieved documents; 2) Classifier Forward Selection (CFS) : we set a judgment condition to each retrieved document in the second-retrieval stage to filter out irrelevant documents which are useless or even play a negative role in the answer. In addition, we will train a classifier $C$ by a small model with millisecond-level runtime to prevent excessive delays in our pipelines. DR-RAG involves a small binary-classification model where the input consists of $\boldsymbol{q}$ and two documents. The training objective is to determine the potential contribution of the documents to answering $\boldsymbol{q}$. The specific settings are as follows:
\begin{align}\footnotesize
\begin{split}
    \text{C}(\boldsymbol{q}, \boldsymbol{d}^*, \boldsymbol{d}^*) & =positive \\
    \text{C}(\boldsymbol{q}, \boldsymbol{d}^*, \boldsymbol{d}^{\Delta})& =negative \\
    \text{C}(\boldsymbol{q}, \boldsymbol{d}^{\Delta}, \boldsymbol{d}^{\Delta}) & = negative \\
\end{split}
\end{align}
where $\text{C}$ represents the classifier. $positive$ and $negative$ indicate whether the two documents are critical for the query.

\begin{table*}[t]
\footnotesize
\caption{Results on different datasets with Llama3-8B as LLM. Adaptive Retrieval and Self-RAG conduct the retrieval module only under specific conditions (unpopular query entities or special retrieval tokens), so their time overhead is much less than other methods. We emphasize our results in bold.}
\label{total result}
\centering
\renewcommand{\arraystretch}{1.1}%调行距
\setlength{\tabcolsep}{2.85pt}
\linespread{1}
\begin{tabular}{@{}lccccccccccccccc@{}}
\toprule
                              & \multicolumn{5}{c}{\textbf{MuSiQue}}                                             & \multicolumn{5}{c}{\textbf{HotpotQA}}                                            & \multicolumn{5}{c}{\textbf{2Wiki}}                                     \\ \cmidrule(l){2-16} 
\textbf{Methods}              & EM             & F1             & Acc            & Step          & Time          & EM             & F1             & Acc            & Step          & Time          & EM             & F1             & Acc            & Step          & Time          \\ \midrule
\textbf{Single-step Approach} & 13.80          & 22.80          & 15.20          & 1.00          & 1.00          & 34.40          & 46.15          & 36.40          & 1.00          & 1.00          & 41.60          & 47.90          & 42.80          & 1.00          & 1.00          \\
\textbf{Adaptive Retrieval}   & 6.40           & 15.80          & 8.00           & 0.50          & 0.55          & 23.60          & 32.22          & 25.00          & 0.50          & 0.55          & 33.20          & 39.44          & 34.20          & 0.50          & 0.55          \\
\textbf{Self-RAG}             & 1.60           & 8.10           & 12.00          & 0.73          & 0.51          & 6.80           & 17.53          & 29.60          & 0.73          & 0.45          & 4.60           & 19.59          & 38.80          & 0.93          & 0.49          \\
\textbf{Adaptive-RAG}         & 23.60          & 31.80          & 26.00          & 3.22          & 6.61          & 42.00          & 53.82          & 44.40          & 3.55          & 5.99          & 40.60          & 49.75          & 44.73          & 2.63          & 4.68          \\
\textbf{Multi-step Approach}  & 23.00          & 31.90          & 25.80          & 3.60          & 7.58          & 44.60          & 56.54          & 47.00          & 5.53          & 9.38          & 49.60          & 58.85          & 55.40          & 4.17          & 7.37          \\
\textbf{DR-RAG(Ours)}         & \textbf{26.97} & \textbf{38.90} & \textbf{34.03} & \textbf{1.00} & \textbf{1.54} & \textbf{48.58} & \textbf{62.87} & \textbf{55.68} & \textbf{1.00} & \textbf{1.40} & \textbf{49.60} & \textbf{55.62} & \textbf{55.18} & \textbf{1.00} & \textbf{1.43} \\ \bottomrule
\end{tabular} 
\vspace{-0.4cm}
\end{table*}

\textbf{Classifier Inverse Selection}\hspace{0.4cm}In this approach, we selectively exclude some irrelevant documents from the retrieved $k$ documents to minimize document redundancy. Specifically, after obtaining $k$ documents in stages, we pair them as <$\boldsymbol{q}, \boldsymbol{d_m}, \boldsymbol{d_n}$> and get $C_k^2$ pairs. The pairs, with the current query $\boldsymbol{q}$, are collectively fed into the classifier $C$. Similarly, when the classification result of a document and the remaining $k$-1 documents is negative, then we consider the document as redundant and should be removed. The whole process is:
\begin{align}\footnotesize
\begin{split}
    Cnt &= Cnt \cup \{\boldsymbol{d}'_{i,j} \mid \boldsymbol{d}'_{i,j} \not\in Cnt \land \text{$first$}\} \\
    P_{i,j} &= 
    \begin{cases} 
    1 & \text{if } \exists i,  C(\boldsymbol{q}, \boldsymbol{d'}_{i,j}, \boldsymbol{d}_i) = positive \\
   0 & \text{$otherwise$}
    \end{cases} \\
     Cnt &= Cnt - \{\boldsymbol{d'}_{i,j} \mid P_{i,j} = 0\}\\
    \text{$answer$} &= \texttt{LLM}(\texttt{Concat}(\boldsymbol{q}, Cnt))
\end{split}
\end{align}
where $\mbox{-}$ represents complement. $Cnt = Cnt - \{\boldsymbol{d'}_{i,j} \mid P_{i,j} = 0\}$ means $\boldsymbol{d'}_{i,j}$ in the second-retrieval stage is classified as $negative$ combined with all $\boldsymbol{di}$ in the first-retrieval stage, then $\boldsymbol{d'}_{i,j}$ will be removed.

% However, given that the classifier's accuracy is not 100 percent, there is an inherent risk of excluding some of the necessary documents $\boldsymbol{d}^*$, from the original set of $k$ documents, consequently leading to a reduction in our document recall rate. In summary, the fundamental concept in this document involves sacrificing a marginal portion of the recall rate to minimize the proportion of irrelevant information provided to the large model.

\textbf{Classifier Forward Selection}\hspace{0.4cm}Unlike the CIS method, CFS method aims to remove the irrelevant dynamic-relevant documents in the second-retrieval stage. To achieve this goal, we search for a document $\boldsymbol{d_{n}}$ from $D$ according to the <$\boldsymbol{q},\boldsymbol{d_{m}}$> pair, and feed both the query and documents into $C$. When the classification result is negative, we will exclude the dynamic-relevant document in the current retrieved documents, and search for the next dynamic-relevant document which can be classified as positive with $_{m}$. The whole process is:
\begin{align}\footnotesize
\begin{split}
    P_{i,j} &=
    \begin{cases} 
    1 & \text{if }  C(\boldsymbol{q}, \boldsymbol{d}_i, \boldsymbol{d}'_{i,j}) = positive \\
    0 & \text{$otherwise$}
    \end{cases}   \\
     Cnt &= Cnt \cup \{\boldsymbol{d}'_{i,j} \mid P_{i,j} = 1 \land first\}\\
    \text{$answer$} &= \texttt{LLM}(\texttt{Concat}(\boldsymbol{q}, Cnt))
\end{split}
\end{align}
where $Cnt = Cnt \cup \{\boldsymbol{d}'_{i,j} \mid P_{i,j} = 1 \land first\}$ means that for a given $\boldsymbol{d_i}$, the first $\boldsymbol{d}'_{i,j}$ in the second-retrieval stage classified as $positive$ combined with $\boldsymbol{d_i}$ will be considered as dynamic-relevant document and placed into $Cnt$.

\section{Experiment Settings}
The experimental details will be described in this section. Due to space constraints, the descriptions of implementation details, retrieval strategy and baseline can seen in Appendix \ref{imple}, \ref{retrieve} and \ref{baseline}. 

\subsection{Dataset}
We verify the effectiveness of our proposed framework on three multi-hop QA datasets, including HotpotQA, 2Wiki and MuSiQue \cite{yang2018hotpotqa, ho2020constructing, trivedi2022MuSiQue}. The datasets require the system to comprehensively collect and contextualize information from multiple documents to answer more complex queries.

\begin{table*}[t]
\footnotesize
\caption{Results on different LLMs and strategies compared to Adaptive-RAG. We set gpt-3.5-turbo and Llama3-8b as the base LLM. We emphasize our best results in bold. Top-k means the total number of retrieved documents.}
\renewcommand{\arraystretch}{1.1}%调行距
\setlength{\tabcolsep}{2.85pt}
\linespread{1}
\centering
\label{gpt3.5 and llama3}
\begin{tabular}{@{}cclccccccccc@{}}
\toprule
\multicolumn{3}{c}{\textbf{}}                                                                                          & \multicolumn{3}{c}{\textbf{MuSiQue}}             & \multicolumn{3}{c}{\textbf{HotpotQA}}            & \multicolumn{3}{c}{\textbf{2Wiki}}     \\ \cmidrule(l){4-12} 
\multicolumn{1}{l}{\textbf{top-$k$}} & \textbf{LLMs}                        & \textbf{Methods}                            & EM             & F1             & Acc            & EM             & F1             & Acc            & EM             & F1             & Acc            \\ \midrule
\multicolumn{1}{l}{\textbf{}}      & \textbf{}                           & \textbf{Adaptive-RAG}                       & 23.60          & 31.80          & 26.00          & 42.00          & 53.82          & 44.00          & 40.60          & 49.75          & 46.40          \\ \midrule
\multirow{6}{*}{\textbf{3}}        & \multirow{3}{*}{\textbf{gpt-3.5}}   & \textbf{Query Document Concatenation} & 21.40          & 32.20          & 29.90          & 37.80          & 51.56          & 52.60          & 36.20          & 48.99          & 51.40          \\
                                   &                                     & \textbf{Classifier Inverse Selection}       & 23.70          & 33.70          & 29.40          & 41.20          & 53.91          & 53.40          & 38.00          & 51.48          & 54.20          \\
                                   &                                     & \textbf{Classifier Forward Selection}       & \textbf{26.00} & \textbf{36.20} & \textbf{35.00} & \textbf{43.80} & \textbf{58.83} & \textbf{55.00} & \textbf{48.40} & \textbf{60.13} & \textbf{64.20} \\ \cmidrule(l){2-12} 
                                   & \multirow{3}{*}{\textbf{Llama3-8B}} & \textbf{Query Document Concatenation} & 21.30          & 32.00          & 27.10          & 43.85          & 56.38          & 49.88          & 40.84          & 48.61          & 43.76          \\
                                   &                                     & \textbf{Classifier Inverse Selection}       & 20.90          & 31.90          & 27.00          & 44.88          & 57.05          & 50.94          & 42.71          & 50.82          & 46.45          \\
                                   &                                     & \textbf{Classifier Forward Selection}       & \textbf{26.50} & \textbf{37.40} & \textbf{32.60} & \textbf{48.71} & \textbf{62.42} & \textbf{55.26} & \textbf{50.48} & \textbf{59.29} & \textbf{56.07} \\ \midrule
\multirow{6}{*}{\textbf{4}}        & \multirow{3}{*}{\textbf{gpt-3.5}}   & \textbf{Query Document Concatenation} & 25.37          & 25.05          & 35.70          & 42.15          & 55.79          & 53.31          & 50.60          & 59.99          & 62.20          \\
                                   &                                     & \textbf{Classifier Inverse Selection}       & 25.80          & 36.50          & 35.60          & 42.00          & 56.10          & 54.36          & 49.40          & 60.40          & 65.00          \\
                                   &                                     & \textbf{Classifier Forward Selection}       & \textbf{25.80} & \textbf{37.60} & \textbf{38.60} & \textbf{45.00} & \textbf{60.55} & \textbf{57.40} & \textbf{52.40} & \textbf{63.95} & \textbf{69.60} \\ \cmidrule(l){2-12} 
                                   & \multirow{3}{*}{\textbf{Llama3-8B}} & \textbf{Query Document Concatenation} & 25.10          & 37.10          & 32.00          & 46.05          & 59.98          & 53.09          & 45.79          & 54.62          & 49.70          \\
                                   &                                     & \textbf{Classifier Inverse Selection}       & 25.70          & 37.50          & 32.70          & 48.02          & 61.43          & 54.36          & \textbf{50.52}          & \textbf{59.54}          & \textbf{55.94}         \\
                                   &                                     & \textbf{Classifier Forward Selection}       & \textbf{27.10} & \textbf{39.30} & \textbf{34.30} & \textbf{48.30} & \textbf{62.81} & \textbf{55.22} & 50.30 & 59.40 & 55.85 \\ \midrule
\multirow{6}{*}{\textbf{6}}        & \multirow{3}{*}{\textbf{gpt-3.5}}   & \textbf{Query Document Concatenation} & 28.80          & 41.30          & 38.60          & 45.41          & 60.09          & 61.60          & 48.20          & 62.58          & 67.00          \\
                                   &                                     & \textbf{Classifier Inverse Selection}       & \textbf{31.20}          & \textbf{42.70}          & 40.20          & 45.80          & 60.38          & 60.80          & \textbf{52.60}          & \textbf{65.93}          & \textbf{71.20}          \\
                                   &                                     & \textbf{Classifier Forward Selection}       & 28.40 & 41.10 & \textbf{40.60} & \textbf{48.20} & \textbf{63.83} & \textbf{63.60} & 49.80 & 63.74 & 68.40 \\ \cmidrule(l){2-12} 
                                   & \multirow{3}{*}{\textbf{Llama3-8B}} & \textbf{Query Document Concatenation} & 25.90          & 38.10          & 33.70          & 46.19          & 60.21          & 54.14          & 44.02          & 53.08          & 48.73          \\
                                   &                                     & \textbf{Classifier Inverse Selection}       & \textbf{27.50}          & 39.40          & 34.60          & 47.34          & 61.30          & 54.55          & \textbf{50.59}          & \textbf{59.58}          & \textbf{56.29}          \\
                                   &                                     & \textbf{Classifier Forward Selection}       & 27.30 & \textbf{40.00} & \textbf{35.20} & \textbf{48.73} & \textbf{63.38} & \textbf{56.56} & 48.02 & 57.17 & 53.63 \\ \bottomrule
\end{tabular}   
\vspace{-0.4cm}
\end{table*}

\begin{table}[t]    
\centering
\footnotesize       
\renewcommand{\arraystretch}{1}%调行距    
\setlength{\tabcolsep}{10pt}    
\linespread{0.1}    
\caption{Ablation study on HotpotQA by Llama3-8B.}
\label{tab:w/o concat}
\begin{tabular}{@{}ccccc@{}}
\toprule
\textbf{top-k} & \textbf{LLMs} & \textbf{EM} & \textbf{F1} & \textbf{Acc} \\ \midrule
\multirow{5.5}{*}{\textbf{3}} & \textbf{CFS} & \textbf{48.71} & \textbf{62.42} & \textbf{55.26} \\ \cmidrule(l){2-5} 
 & \textbf{w/o QDC} & 46.26 & 59.37 & 52.60 \\ \cmidrule(l){2-5} 
 & \textbf{CIS} & \textbf{44.88} & \textbf{57.05} & \textbf{50.94} \\ \cmidrule(l){2-5} 
 & \textbf{w/o QDC} & 44.82 & 57.00 & 50.88 \\ \midrule
\multirow{5.5}{*}{\textbf{4}} & \textbf{CFS} & \textbf{48.30} & \textbf{62.81} & \textbf{55.22} \\ \cmidrule(l){2-5} 
 & \textbf{w/o QDC} & 47.78 & 61.63 & 54.73 \\ \cmidrule(l){2-5} 
 & \textbf{CIS} & \textbf{48.02} & \textbf{61.43} & \textbf{54.36} \\ \cmidrule(l){2-5} 
 & \textbf{w/o QDC} & 46.54 & 59.28 & 52.80 \\ \midrule
\multirow{5.5}{*}{\textbf{6}} & \textbf{CFS} & \textbf{48.73} & \textbf{63.38} & \textbf{56.56} \\ \cmidrule(l){2-5} 
 & \textbf{w/o QDC} & 48.00 & 63.10 & 56.29 \\ \cmidrule(l){2-5} 
 & \textbf{CIS} & \textbf{47.34} & \textbf{61.30} & \textbf{54.55} \\ \cmidrule(l){2-5} 
 & \textbf{w/o QDC} & 46.30 & 60.00 & 53.95 \\ \bottomrule
\end{tabular}
\vspace{-0.4cm}
\end{table}

\section{Results and Analysis}

\begin{table*}[t]
\footnotesize
\caption{Recall rate and actual numbers under different retrieval strategies. Actual numbers represents the actual numbers of documents that we feed into LLMs. A smaller number means fewer redundant documents.}
\renewcommand{\arraystretch}{1.1}%调行距
\setlength{\tabcolsep}{1.5pt}
\linespread{1}
\centering
\label{diffenent strategy results}
\begin{tabular}{@{}clcccccc@{}}
\toprule
\multicolumn{1}{l}{}        &                                             & \multicolumn{2}{c}{\textbf{MuSiQue}}                               & \multicolumn{2}{c}{\textbf{HotpotQA}}                              & \multicolumn{2}{c}{\textbf{2Wiki}}                       \\ \cmidrule(l){3-8} 
\textbf{top-$k$}              & \textbf{Retrieval Strategies}                  & \multicolumn{1}{l}{Recall rate} & \multicolumn{1}{l}{Actual numbers} & \multicolumn{1}{l}{Recall rate} & \multicolumn{1}{l}{Actual numbers} & \multicolumn{1}{l}{Recall rate} & \multicolumn{1}{l}{Actual numbers} \\ \midrule
\multirow{5}{*}{\textbf{3}} & \textbf{BM25}                               & 37.57                           & 3.00                             & 64.67                           & 3.00                             & 57.46                           & 3.00                             \\
                            & \textbf{Similarity Matching}               & 58.31                           & 3.00                             & 80.33                           & 3.00                             & 74.34                           & 3.00                             \\
                            & \textbf{Query Document Concatenation} & 58.44                           & 3.00                             & 86.12                           & 3.00                             & 74.48                           & 3.00                             \\
                            & \textbf{Classifier Inverse Selection}       & 57.42                           & \textbf{2.82}                            & 79.80                           & \textbf{2.74}                             & 74.34                           & \textbf{2.60}                             \\
                            & \textbf{Classifier Forward Selection}       & \textbf{66.45}                           & 2.95                             & \textbf{88.89}                           & 2.89                             & \textbf{87.81}                           & 2.83                             \\ \midrule
\multirow{5}{*}{\textbf{4}} & \textbf{BM25}                               & 43.52                           & 4.00                             & 70.72                           & 4.00                             & 63.95                           & 4.00                             \\
                            & \textbf{Similarity Matching}               & 66.20                           & 4.00                             & 85.57                           & 4.00                             & 80.05                           & 4.00                             \\
                            & \textbf{Query Document Concatenation} & 70.30                           & 4.00                             & 90.20                           & 4.00                             & 89.89                           & 4.00                             \\
                            & \textbf{Classifier Inverse Selection}       & 66.45                           & \textbf{3.61}                             & 88.37                           & \textbf{3.43}                             & 89.71                           & \textbf{3.03}                             \\
                            & \textbf{Classifier Forward Selection}       & \textbf{73.83}                           & 3.82                             & \textbf{92.28}                           & 3.75                             & \textbf{94.13}                           & 3.41                             \\ \midrule
\multirow{5}{*}{\textbf{6}} & \textbf{BM25}                               & 51.69                           & 6.00                             & 79.93                           & 6.00                             & 73.95                           & 6.00                             \\
                            & \textbf{Similarity Matching}               & 76.78                           & 6.00                             & 92.59                           & 6.00                             & 88.83                           & 6.00                             \\
                            & \textbf{Query Document Concatenation} & 79.46                           & 6.00                             & 94.00                           & 6.00                             & 95.04                           & 6.00                             \\
                            & \textbf{Classifier Inverse Selection}       & 77.13                           & \textbf{5.23}                             & 93.24                           & \textbf{4.72}                             & 94.79                           & \textbf{3.72}                             \\
                            & \textbf{Classifier Forward Selection}       & \textbf{83.01}                           & 5.69                             & \textbf{96.27}                           & 5.35                             & \textbf{98.04}                           & 4.63                             \\ \bottomrule
\end{tabular}
\vspace{-0.4cm}
\end{table*}

\subsection{Main Results}

Table \ref{total result} and \ref{gpt3.5 and llama3} present the performance of DR-RAG in answering multi-hop query, and highlight the advantages of our approach compared to the sota RAG framework \cite{jeong2024adaptive, asai2024selfrag} across multiple metrics, which is in line with our expectations. Table \ref{diffenent strategy results} shows the performance of DR-RAG across various retrieval strategies.

As shown in Table \ref{total result}, when retrieving the same $k$ documents, DR-RAG can achieve a higher recall rate and a higher percentage of correct answers. From the results, DR-RAG achieves better performance than other baseline RAG frameworks (self-RAG and Adaptive-RAG) on all three metrics. Moreover, DR-RAG is also less than other RAG frameworks in terms of the number of LLMs responses and the time consumed in QA systems.

\subsection{Analysis}

\textbf{Ablation Study}\hspace{0.4cm}We propose a two-stage retrieval and classifier selection strategies to mine the dynamic relevance of documents. As shown in Table \ref{gpt3.5 and llama3}, we apple two classification methods based on QDC, and the experimental results have achieved further improvement. Table \ref{tab:w/o concat} shows the comparison of the effect of DR-RAG with and without QDC. Quantitatively, CIS and CFS can improve DR-RAG's performance by 2.3\% and 4.7\% on Acc metric against QDC, while DR-RAG reduces performance by 1.1\% and 0.7\% on Acc metric without QDC. The results demonstrate that the two strategies are able to efficiently extract document relevance and achieve more accurate answers.

\begin{table}[t]\footnotesize
\caption{Results of different classifier on HotpotQA dataset as Llama3-8B.}
\renewcommand{\arraystretch}{1.3}%调行距
\setlength{\tabcolsep}{2pt}
\linespread{1}
\centering
\label{tab:classifier}
\begin{tabular}{ccccc}
\hline
\textbf{top-$k$} & \textbf{Classifier} & \textbf{EM} & \textbf{F1} & \textbf{Acc} \\ \hline
\multirow{3}{*}{\textbf{3}} & \textbf{Bigbird-base(125M)} & 48.71 & 62.42 & 55.26 \\ \cline{2-5} 
 & \textbf{Bigbird-large(355M) }& \textbf{49.20} & \textbf{62.89} & \textbf{55.55} \\ \cline{2-5} 
 & \textbf{Longformer(147M)} & 49.00 & 62.80 & 55.34 \\ \hline
\multirow{3}{*}{\textbf{4}} & \textbf{Bigbird-base(125M)} & 48.30 & 62.81 & 55.22 \\ \cline{2-5} 
 & \textbf{Bigbird-large(355M) }& \textbf{49.12} & \textbf{63.70} & \textbf{56.00} \\ \cline{2-5} 
 & \textbf{Longformer(147M)} & 50.27 & 64.41 & 56.90 \\ \hline \multirow{3}{*}
{\textbf{6}} & \textbf{\textbf{Bigbird-base(125M)}} & 48.73 & 63.38 & 56.56 \\ \cline{2-5} 
 & \textbf{Bigbird-large(355M) }& \textbf{49.02} & \textbf{63.58} & \textbf{56.52} \\ \cline{2-5} 
 & \textbf{Longformer(147M)} & 50.08 & 64.69 & 57.30 \\ \hline
\end{tabular}
\vspace{-0.4cm}
\end{table}

\begin{table}[t]\footnotesize 
\centering
\caption{Results of 500 samples sampled on HotpotQA dataset based on gpt-3.5-turbo and gpt-4-turbo.}
\label{tab:gpt-3-4-5}
\setlength{\tabcolsep}{1pt}    
\linespread{1}    
\renewcommand{\arraystretch}{1}%调行距    
\begin{tabular}{@{}ccccc@{}}
\toprule
\textbf{top-$k$} & \textbf{LLMs} & \textbf{EM} & \textbf{F1} & \textbf{Acc} \\ \midrule
\multirow{2.5}{*}{\textbf{3}} & \textbf{gpt-3.5-turbo} & 43.80 & 58.83  & 55.00 \\ \cmidrule(l){2-5} 
 & \textbf{gpt-4-turbo} & \textbf{51.60} & \textbf{68.93} & \textbf{67.20} \\ \midrule
\multirow{2.5}{*}{\textbf{4}} & \textbf{gpt-3.5-turbo} & 45.00 & 60.55 & 57.40  \\ \cmidrule(l){2-5} 
 & \textbf{gpt-4-turbo} & \textbf{53.40} & \textbf{70.75} & \textbf{69.80} \\ \midrule
\multirow{2.5}{*}{\textbf{6}} & \textbf{gpt-3.5-turbo} & 48.20 & 63.83 & 63.60  \\ \cmidrule(l){2-5} 
 & \textbf{gpt-4-turbo} & \textbf{59.20} & \textbf{75.43} & \textbf{77.20} \\ \bottomrule
\end{tabular}
\vspace{-0.4cm}
\end{table}

\textbf{Effects of Classifier and LLM}\hspace{0.4cm}Compared to gpt-3.5-turbo, gpt-4-turbo with better document comprehension has the ability to accurately capture the critical information to answer a query. As for textual responses, gpt-4-turbo generates responses of higher quality and more accurate content. Quantitatively, as shown in Table \ref{tab:gpt-3-4-5}, gpt-4-turbo improve by an average of 9.07\%, 10.63\%, and 12.73\% against gpt-3.5-turbo on three metrics. As shown in Table \ref{tab:classifier}, when switching to different kinds or sizes of classifiers, the difference in the metrics is negligible (the extreme difference of EM, F1, and Acc is less than 2\%), which suggests that our approach is applicable to different classifiers and that the classifier has little impact on our framework.

\textbf{Effects of Recall Rate}\hspace{0.4cm}The ability of LLMs to answer domain-specific query correctly almost depends on whether all the necessary information is included in the prompt context. When relevant information is missing, it is difficult for LLMs with the hallucination problem to accurately answer the query. Table \ref{some examples} illustrates the answers of the query with and without sufficient information provided to LLMs. As seen in Table \ref{diffenent strategy results}, in 2Wiki, our retrieval strategy already achieves a recall rate of 98\% when top-$k$ is 6. When we feed enough relevant information into LLMs, the accuracy of their answers can be improved accordingly. CFS method achieves higher recall rate by 26.4\% and 8.6\% than BM25 and SM methods, respectively, which proves the feasibility of DR-RAG.

\begin{table*}[t]    
\caption{Case study with Llama3-8B, where we present the factual error in \textcolor{red}{red} and the accurate infomation in \textcolor{blue}{blue}.}
\centering
\footnotesize    
\label{some examples}    
\renewcommand{\arraystretch}{1}%调行距    
\setlength{\tabcolsep}{1.8pt}    
\linespread{1}    
\begin{tabular}{@{}llll@{}}
\toprule
\textbf{Datasets} & \textbf{Queries} & \textbf{Traditional RAG} & \textbf{DR-RAG (Ours)} \\ \midrule
\textbf{MuSiQue} &
  \begin{tabular}[t]{@{}l@{}}Who is the spouse of \\ the Green performer?\end{tabular} &
  {\color[HTML]{333333} \begin{tabular}[t]{@{}l@{}}\textbf{Retrieval strategy}: SM\\ \textbf{Reasoning}: The spouse of the Green \\ performer is \textcolor{red}{Debbie Allen} because \\ she is the \textcolor{red}{sister} of the Green perfor-\\ mer's sister.\\ The answer is: \textcolor{red}{Debbie Allen}.\end{tabular}} &
  {\color[HTML]{333333} \begin{tabular}[t]{@{}l@{}}\textbf{Retrieval strategy}: CFS \\ \textbf{Reasoning}: The Green performer mentioned \\ in the question is Steve Hillage. Then we i-\\ dentify that  \textcolor{blue}{Miquette Giraudy} is mentioned \\ as \textcolor{blue}{his partner}. \\ The answer is: \textcolor{blue}{Miquette Giraudy}.\end{tabular}} \\ \midrule
\textbf{HotpotQA} &
  \begin{tabular}[t]{@{}l@{}}what year was the cover \\ artist of Multiverse: Ex-\\ ploring Poul Anderson's\\ Worlds born ?\end{tabular} &
  \begin{tabular}[t]{@{}l@{}}\textbf{Retrieval strategy}: SM\\ \textbf{Reasoning}: The book “Multiverse:\\  Exploring Poul Anderson's Worlds” \\ was first published in hardcover in \\ May 2014 by Subterranean Press.\\ And assuming Bob Eggleton was \\ around 50 years old. We can cal-\\ culate his birth year is \textcolor{red}{1964}.\\ The answer is: \textcolor{red}{1964}.\end{tabular} &
  \begin{tabular}[t]{@{}l@{}}\textbf{Retrieval strategy}: CFS\\ \textbf{Reasoning}: the cover artist of “Multiverse: \\ Exploring Poul Anderson's Worlds” is Bob \\ Eggleton. We also know that Bob Eggleton \\ was born on September 13, \textcolor{blue}{1960}.\\ The answer is: \textcolor{blue}{1960}\end{tabular} \\ \midrule
\textbf{2Wiki} &
  \begin{tabular}[t]{@{}l@{}}What nationality is the \\ director of film Downpour \\ (Film)?\end{tabular} &
  \begin{tabular}[t]{@{}l@{}}\textbf{Retrieval strategy}: SM\\ \textbf{Reasoning}: we know that “Down-\\ pour” is directed by Bahram Bayzai.\\ the nationality of Bahram Bayzai is \\ not explicitly mentioned in the pro-\\ vided contexts. However, based on \\ my knowledge, Bahram Bayzai is \\ an \textcolor{red}{Iranian} director.\\ The answer is: \textcolor{red}{Iranian}\end{tabular} &
  \begin{tabular}[t]{@{}l@{}}\textbf{Retrieval strategy}: CFS\\ \textbf{Reasoning}: It is mentioned that the film \\ “Downpour” is directed by Bahram Bayzai.\\ And Bahram Beyzaie is associated with \\ \textcolor{blue}{persian} letters, arts. So the director of the \\ film “Downpour” is of \textcolor{blue}{persia} nationality.\\ The answer is \textcolor{blue}{persia}.\end{tabular} \\ \bottomrule
\end{tabular}
\vspace{-0.4cm}
\end{table*}

\textbf{Effects of Redundant Information}\hspace{0.4cm}We hypothesise that if there is less redundant information in the contextual knowledge, LLMs can fully understand the query to reduce the hallucination. Therefore, CIS method is devised to validate this hypothesis. Invalid information may increase by about 30\% as the number of documents fed into LLMs increases, but LLMs fail to judge the information when answering. LLMs may refer to redundant information and provide an answer with incorrect information. The results all validate our hypothesis that we should provide LLMs as little redundant or incorrect information as possible throughout the RAG process. CIS method is effective in removing redundant information, but it may reduce the quality of responses when the reduction in recall is too large. Even though we feed all the relevant documents into LLMs, it is still possible to fail to get the right answer. In Table \ref{diffenent strategy results}, on dataset 2Wiki, when the number of documents $k$ provided to LLMs at 4 and 6, there is only a slight increase from CIS to CFS in the recall and instead a decrease in the metrics. Therefore, CFS method is propsed to balance redundant and relevant information.

\textbf{Increase Recall with Lower Documents}\hspace{0.4cm}In CFS method, it seems impossible to find a match for every <$\boldsymbol{q},\boldsymbol{d}$> pair in the second-retrieval stage because the documents we need have been retrieved. Therefore, there will be cases where the total number of our retrieved documents is less than $k$. For instance, in the HotpotQA dataset, when $k$ is set to 6, the average number of documents actually provided to LLMs is 5.35, thereby reducing irrelevant information to some extent. CFS method in Table \ref{diffenent strategy results} achieves a higher recall rate while retrieving fewer actual numbers of documents compared to QDC method. CFS method yields higher scores across the three metrics in our experiments and achieves more significant retrieval capabilities with lower redundant inputs than other methods.

\begin{figure}[t]
    \centering
    \includegraphics[width=1\linewidth]{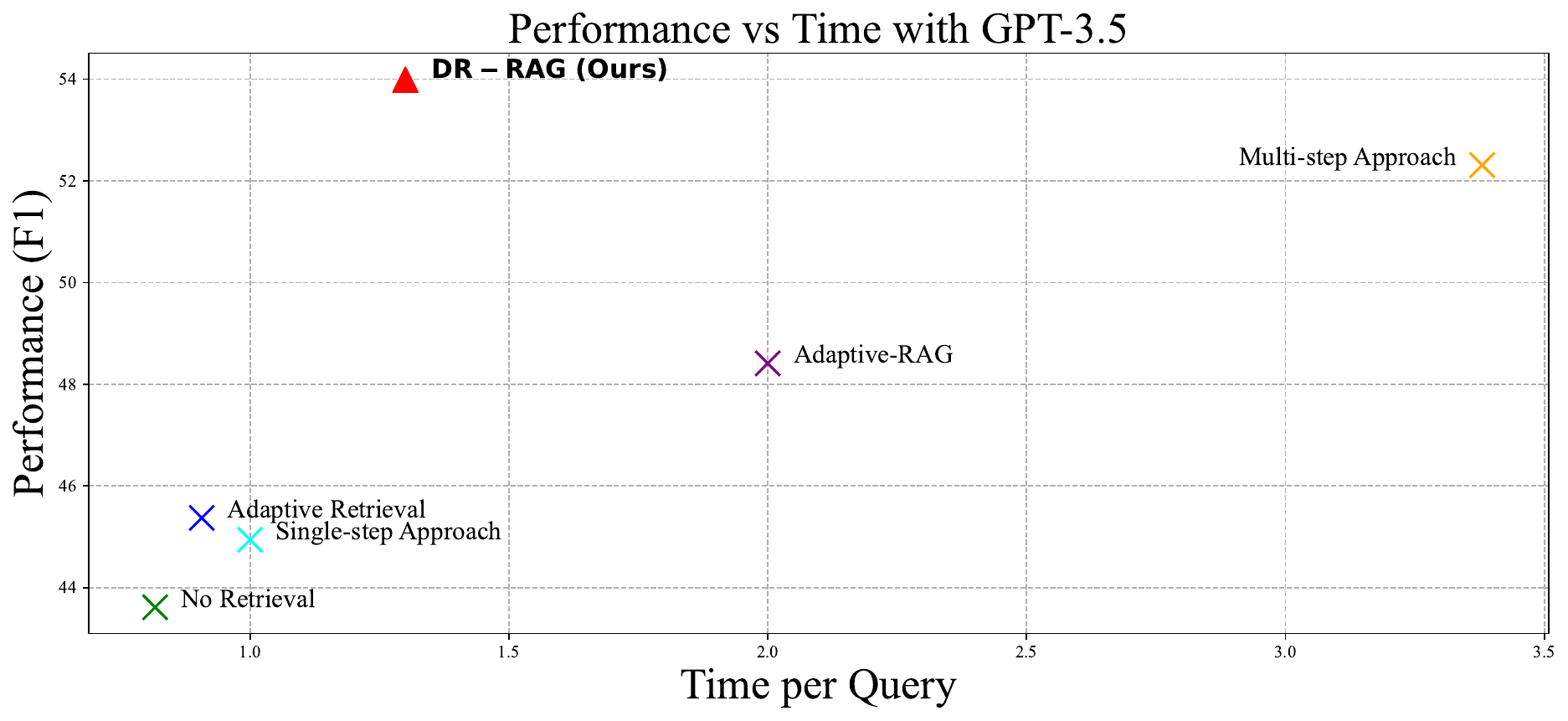}
        % \caption{Caption 1}
    \caption{QA performance (F1) and time for different RAG frameworks. We use the GPT-3.5-turbo as the base LLM on the multi-hop QA datasets (MuSiQue, HotpotQA and 2Wiki).}
    \label{fig:enter-label}
    \vspace{-0.4cm}
\end{figure}

\textbf{Time for One Response}\hspace{0.4cm}Compared to previous RAG frameworks, DR-RAG also achieves better time optimization during the whole process. Other RAG frameworks may call LLMs multiple times, resulting in high computational cost. In fact, the inference time of LLMs is also a worthwhile optimization in the applications. It takes a lot of time to call LLMs once, and calling them multiple times presents a catastrophic challenge in terms of time overhead. Therefore, we attempt to design a small model with relatively few parameters to achieve better optimization rather than calling LLMs multiple times. In Fig. \ref{fig:enter-label} and Table \ref{total result}, compared to Adaptive-RAG, we have achieved an average 74.2\% reduction in time overheads. Therefore, we can conclude that we can achieve better experimental efficiency and the time overhead makes DR-RAG valuable in applications.

\textbf{Case Study}\hspace{0.4cm}We conduct a case study to qualitatively compare our DR-RAG against the traditional RAG. Table \ref{some examples} demonstrates the specific inference cases on the multi-hop datasets. For example, in MuSiQue dataset, our DR-RAG identifies the answer to the query by only using the LLM's parametric knowledge about ‘partner’. Traditional RAG sometimes generate incorrect responses due to the inclusion of irrelevant information about ‘sister’. Meanwhile, faced with a complex query, DR-RAG can first retrieve static-relevant documents based on ‘cover artist’ and ‘Multiverse: Exploring Poul Anderson's Worlds’ to get the name ‘Bob Eggleton’. Then, in the second-retrieval stage, by combining the name ‘Bob Eggleton’ with ‘born’ in the query, dynamic-relevant documents can be retrieved to obtain the answer ‘1960’.

\section{Conclusion}
This paper presents DR-RAG, an innovative RAG framework designed to enhance document retrieval accuracy by leveraging the relevance of different documents in various QA scenarios. Throughout this research, we explore diverse retrieval strategies and conduct comprehensive experimental comparisons. Ultimately, we adopt CFS as the final framework, which not only reduces the number of redundant document but also achieves the most superior performance. Additionally, we analyze the utilization of dynamic document relevance under constrained training resources. The experimental results demonstrate that DR-RAG significantly improves answer quality and reduces the time required for QA systems.

\section{Limitations}
While DR-RAG has demonstrated excellent performance across multiple datasets for multi-hop QA, its implementation requires the prior training of a distinct classifier. It is uncertain whether our classifier will be effective with niche domains. Therefore, DR-RAG can serve as an invaluable inspiration to train a classifier with private data. In the future, we will collect more comprehensive data to train a more applicable classifier for various QA tasks.

% Please add the following required packages to your document preamble:
% \usepackage{booktabs}
% \usepackage{multirow}
% \usepackage[table,xcdraw]{xcolor}
% Beamer presentation requires \usepackage{colortbl} instead of \usepackage[table,xcdraw]{xcolor}

\section{Ethics Statement}
DR-RAG substantiates its efficacy in real-world scenarios, which are characterized by diverse user queries. However, given the potential variability in user inputs, which may span a range from benign to offensive, it is imperative to consider scenarios where inputs might be detrimental. Such instances could facilitate the retrieval of objectionable content and lead to unsuitable responses by retrieval-augmented LLMs. Addressing this concern necessitates the development of robust methodologies to detect and mitigate offensive or inappropriate content in both user inputs and the documents retrieved within the RAG framework. This area represents a critical part for future research.

% Bibliography entries for the entire Anthology, followed by custom entries
%\bibliography{anthology,custom}
% Custom bibliography entries only

\bibliography{latex/acl_latex}

\nocite{ke2024bridging}
\nocite{luo2023divide}
\nocite{wang2023learning}
\nocite{yu2023augmentation}
\nocite{trivedi2022interleaving}
\nocite{ma2023query}
\nocite{shao2023enhancing}
\nocite{sun2022recitation}
\nocite{feng2024retrieval}
\nocite{ren2023investigating}
\nocite{wang2024knowledge}
\nocite{xu2023search}
\nocite{chen2023dense}
\nocite{liu2023exploring}
\nocite{cheng2023uprise}
\nocite{yu2022generate}

\appendix

\section{Appendix}
\label{sec:appendix}

\subsection{Implementation Details}\label{imple}
We follow the standard evaluation approach \cite{jeong2024adaptive} and validate our DR-RAG for QA systems by multiple metrics including F1, EM, and Accuracy (Acc). These metrics provide an objective measure between the prediction results and ground truth. In addition, the efficiency is also another issue we have to tackle. Most existing RAG frameworks \cite{asai2024selfrag, jeong2024adaptive} require multiple calls to LLMs for inference. Therefore, we consider the number of inferences by LLMs and the time required for responses as our evaluation. To eliminate the effects of different LLMs, we select gpt-3.5-turbo \cite{achiam2023gpt, brown2020language} and Llama3-8B \cite{llama3modelcard} as base LLMs, and accurately acquire the answers to query based on retrieval documents. For the classifier $C$, we fine-tune bigbird-roberta-base \cite{zaheer2021big} by the entire training set to accommodate longer input tokens. Due to the imbalance between positive and negative samples in the datasets, we sample the positive and negative examples to construct the datasets with the ratio of 1:1. In addition, we sample about 2300 pieces of data in each dataset, which exceeds the existing experiment \cite{jeong2024adaptive} in sample numbers.

\subsection{Retrieval Strategy}\label{retrieve}
DR-RAG aims to solve the problem of low recall in document retrieval. Therefore, five different retrieval strategies are designed to verify the effectiveness of our proposed DR-RAG.
\begin{itemize}
\item BM25: A method to measure the relevance between $\boldsymbol{q}$ and the documents.
\item SM: The retrieval documents will be embedded and stored in $D$ and the similarity between $\boldsymbol{q}$ and $D$ is calculated to extract the $k$ most relevant documents.
\item QDC: We first retrieve $k_1$ documents from $D$, concatenate $\boldsymbol{q}$ with the documents to form multiple pairs and retrieve the $k_2$ most relevant documents for <$\boldsymbol{q},\boldsymbol{d}$> pairs, respectively, until the number of retrieved documents = $k$.
\item CIS: To minimize document redundancy in QDC, all $k$ documents retrieved are pairwise combined, concatenated with $\boldsymbol{q}$ and then fed into $C$ to filter out irrelevant documents.
\item CFS: To remove irrelevant dynamic-relevant documents, after retrieving $k_1$ documents, <$\boldsymbol{q},\boldsymbol{d}$> pairs are matched one by one with the remaining documents for similarity. Simultaneously, they have been fed into $C$ for classification. If classified as negative, the process have been extended to the next document. Otherwise, positive instances will be included in the document set.
\end{itemize}

\subsection{Baseline}\label{baseline}
We conduct a comprehensive comparison of our retrieval strategies against other RAG frameworks. In DR-RAG, we calculate the recall with different retrieval strategies and then evaluate the accuracy of the answers. Therefore, we select BM25 and SM methods \cite{lewis2020retrieval, chan2024rq} as baselines. Moreover, we choose self-RAG \cite{asai2024selfrag} and Adaptive-RAG \cite{jeong2024adaptive}, which are effective RAG frameworks for multi-hop QA, to validate the performance of our DR-RAG. In addition, we add the experimental results of Non-retrieval, original RAG and multi-step approach \cite{trivedi-etal-2023-interleaving} to enrich our comparisons.

\subsection{Related Works}
\subsubsection{RAG for Multi-hop QA}
RAG is a popular framework for LLMs and has received much attention to many tasks, such as QA systems. RAG \cite{lewis2020retrieval} combined a sequence-to-sequence model with external knowledge bases to significantly improve the quality of quizzing and summarization tasks. The decomposition of a complex query \cite{khattab2022demonstrate, press2022measuring, pereira2023visconde, khot2022decomposed, zhou2022least} into a series of simpler sub-queries might inevitably require multiple calls to LLMs, resulting in high computational cost. Adaptive-RAG \cite{jeong2024adaptive} evaluated the complexity of the problem by a classifier and selects the most appropriate retrieval strategy based on the classification results. RQ-RAG \cite{chan2024rq} aimed to improve the performance of models by optimising search query, including rewriting, decomposition and disambiguation. However, it would be inefficient to access LLMs multiple times for each query and unreliable to retrieve all dynamic-relevant documents by a single query.
\subsubsection{Retriever in RAG}
The retriever in a RAG system is the key to verify how the retriever can obtain revelant instant contexts from external knowledge bases and alleviate the hallucination of LLMs. \citet{fan2021augmenting} combined $K$ Nearest Neighbor (KNN) retrieval with a traditional transformer model to dynamically access historical data and provide enough information by a composite memory. \citet{cheng2024lift} proposed Selfmem to make the generated text more relevant to the retrieved information through a self-memory mechanism. Recent research has highlighted the potential applications of LLMs, which can be considered as supervised signals for training retrieval components, even as retrieval components. These findings provide us with new avenues for exploring the ability of retrievers to improve the efficiency of information retrieval based on the document relevance. In our work, we retrieve multiple relevant documents based on the query by a two-stage strategy and design a classifier to determine whether the documents can answer the query, and the remaining relevant documents are fed into LLMs with the query to obtain the answer.

\subsection{More Analysis}
\textbf{Optimization under Resource Constraints}\hspace{0.4cm}The classifier $C$ of document relevance requires certain hardware conditions and resources for data annotation. However, QDC method indicates that dynamic documents relevance can still be utilized without $C$. As seen in Table \ref{diffenent strategy results}, compared to SM method, across all datasets, when top-$k$ is 4 or 6, there is a significant increase of the retrieval recall by 3.84\%. Yet, when top-$k$ is 3, there is also 6\% increase on HotpotQA and a slight increase on the other two datasets. This suggests that by making reasonable choices about top-$k$, even in cases where resources are limited, the performance of retrieval can be optimized by leveraging the relevance of relevant documents, thus improving LLMs' performance in QA tasks.

\textbf{More Cases}\hspace{0.4cm}Table \ref{prompt} shows the prompts we provide to LLMs. \textcolor{purple}{Contexts} contain the documents (Document $i$) after retrieving and selecting. Moreover, we show the case of the classifier for selection in CIS and CFS methods in Table \ref{some examples1}, the output cases compared to Adaptive-RAG in Table \ref{some examples_compared_with_adaptive}, the case of documents retrieved by QDC and CFS methods in Table \ref{some examples_dqc_cfs}, the case of documents retrieved by QDC and CIS methods in Table \ref{some examples_classifier}.

\begin{algorithm*}[tbp]
  \setstretch{1.5}
  \caption{Classifier Forward Selection (CFS)}\label{algo:classifier-forward-selection}
  \begin{algorithmic}[1]
    \Require  
    \Statex Classifier $C$
    \Statex Retrieval Function \texttt{Retriever}
    % \Input  
    \Statex Input query \(\boldsymbol{q}\)
    % \Output  
    \Statex Generated response \( \text{$answer$} \)
    \State Initialize empty context: \( \text{$Cnt$} = \{\} \)
    \State Retrieve $k_1$ documents: \( \{\boldsymbol{d}_1, \boldsymbol{d}_2, \ldots, \boldsymbol{d}_{k_1}\} = \texttt{Retriever}(\boldsymbol{q}) \)
    \State Update context: \( \text{$Cnt$} = \text{$Cnt$} \cup \{\boldsymbol{d}_1, \boldsymbol{d}_2, \ldots, \boldsymbol{d}_{k_1}\} \)
    \For{\(i = 1 \) to \( k_1 \)}
        \State Construct a new query: \( \boldsymbol{{q_i}^*} = \texttt{concat}(\boldsymbol{q}, \boldsymbol{d_i}) \)
    \EndFor
    \State Retrieve full set of documents for each new query: \\
    \( \{\boldsymbol{d}'_{i,1}, \boldsymbol{d}'_{i,2},, \ldots, \boldsymbol{d}'_{i,k_2}\} = \texttt{Retriever}(\boldsymbol{q}_i^*), \text{ for } i = 1, 2, \ldots, k_1 \)
    \For{\(i = 1\) to \(k_1\)}
        \For{\(j = 1\) to \(k_2\)}
            \If{\(\boldsymbol{d}'_{i,j} \not\in \text{$Cnt$} \) and \( C\text(\boldsymbol{q}, \boldsymbol{d}_i, \boldsymbol{d}'_{i, j}) = positive\)}
                \State Update context: \( \text{$Cnt$} = \text{$Cnt$} \cup \{\boldsymbol{d}'_{i,j}\} \)
            \EndIf
        \EndFor
    \EndFor
    \State Combine the input question with the updated context: \( \text{$input$} = \texttt{concat}(\boldsymbol{q}, \text{$Cnt$}) \)
    \State Generate the answer using a large language model: \( \text{$answer$} = \texttt{LLM}(\text{$input$}) \)
    \State \textbf{return} \( \text{$answer$} \)
  \end{algorithmic} 
\end{algorithm*}

\begin{table*}[t]    
\caption{A case of our prompt provided to LLMs.}    
\centering
 \fontsize{8.7}{8.7}\selectfont
\label{prompt}    
\renewcommand{\arraystretch}{1}%调行距    
\setlength{\tabcolsep}{0pt}    
\linespread{1}    
\begin{tabular}{@{}l@{}}
\toprule
\begin{tabular}[t]{@{}l@{}}You are a reading comprehension expert, and you need to complete a reading comprehension task.\\ ------------------------------------------\\ \textcolor{purple}{Contexts}\\ \\ \textbf{Document 1}:\\ Walk, Don't Run is a 1966 Technicolor comedy film directed by Charles Walters and starring Cary Grant in his final film role,\\ Samantha Eggar, and Jim Hutton. The film is a remake of the 1943 film "The More the Merrier" and is set during the Olympic \\ Games\\ \\ \textbf{Document 2}:\\ Douglas Sirk( born Hans Detlef Sierck; 26 April 1897 – 14 January 1987) was a German film director best known for his work\\  in Hollywood melodramas of the 1950s. Sirk started his career in Germany as a stage and screen director, but he left to Holly-\\ wood in 1937 because his Jewish wife was persecuted by the Nazis. In the 1950s, he achieved his greatest commercial success \\ with film melodramas like "Imitation of Life All That Heaven Allows Written on the WindMagnificent Obsession" and "A\\Time to Love and a Time to Die". While those films were initially panned by critics as sentimental women's pictures, they are \\today widely regarded by film directors, critics and scholars as masterpieces. His work is seen as "critique of the bourgeoisie\\in general and of 1950s America in particular", while painting a" compassionate portrait of characters trapped by social con-\\ditions". Beyond the surface of the film, Sirk worked with complex mise enscenes and lush Technicolor colors to subtly un-\\derline his message.\\ \\ \textbf{Document 3}:\\ The Mall, The Merrier is a 2019 Philippine musical family comedy film directed by Barry Gonzales, starring Vice Ganda and \\ Anne Curtis. The film is co-produced by Star Cinema and Viva Films under the working title" Momalland". The film pre-\\miered in Philippine cinemas on December 25, 2019 as one of the official entries to the 2019 Metro Manila Film Festival. "\\The Mall, The Merrier" marks the first on- screen collaboration between Anne Curtis and Vice Ganda, both of whom are\\regular hosts in the noontime variety show" It's Showtime".\\ \\ \textbf{\textbf{Document 4:}}:\\ Robert Wallace Russell( January 19, 1912 – February 11, 1992) was an American writer for movies, plays, and documentaries. \\ He was nominated for two Academy Awards for Best Writing, Original Story and Best Writing, Screenplay on the 1943 film \\ "The More the Merrier". He died in 1992 in New York City, shortly after his 80th birthday.\\ \\ \textbf{Document 5}:\\ Sleep, My Love is a 1948 American film noir directed by Douglas Sirk and starring Claudette Colbert, Robert Cummings and \\ Don Ameche.\\ \\ \textbf{Document 6}:\\ The More the Merrier is a 1943 American comedy film made by Columbia Pictures which makes fun of the housing shortage \\ during World War II, especially in Washington, D.C. The picture stars Jean Arthur, Joel McCrea and Charles Coburn. The\\movie was directed by George Stevens. The film was written by Richard Flournoy, Lewis R. Foster, Frank Ross, and Robert\\ Russell, from" Two's a Crowd", an original story by Garson Kanin( uncredited). This film was remade in 1966 as" Walk,\\Don't Run",  with Cary Grant, Samantha Eggar and Jim Hutton.\\ ------------------------------------------\\ After reading the documents above, answering the following question. Reasoning step by step. At last, you should output the\\final result via the following format:\\ Answer: \textless{}your answer based on the documents\textgreater{};\\ Please answer the question directly.\\ ------------------------------------------\\ \textbf{Question}\\ Which film has the director who died later, The More The Merrier or Sleep, My Love?\\ ------------------------------------------\\ Give your analysis process first, and then output your answer in a specified format.\end{tabular} \\ \bottomrule
\end{tabular}
\end{table*}

\begin{table*}[t]    
\caption{Case of the classifier we train for selection in CIS and CFS methods. We mark relevant information that can influence classification results in \textcolor{blue}{blue}.}    
\centering
 \fontsize{8.5}{8.5}\selectfont
\label{some examples1}    
\renewcommand{\arraystretch}{1}%调行距    
\setlength{\tabcolsep}{1.8pt}    
\linespread{1}    
\begin{tabular}{@{}lll@{}}
\toprule
\textbf{Dataset} &
  \textbf{classified as $positive$} &
  \textbf{classified as $negative$} \\ \midrule
\textbf{MuSiQue} &
  \begin{tabular}[t]{@{}l@{}}\textbf{Query}\\ In \textcolor{blue}{True Grit}, who did the \textcolor{blue}{star play}?\\ \\ \textbf{Document 1}\\ \textcolor{blue}{True Grit} is a 1969 American western film. It is the first\\  film adaptation of Charles Portis' 1968 novel of the same\\  name. The screenplay was written by Marguerite Roberts.\\  The film was directed by Henry Hathaway and \textcolor{blue}{starred} \\ \textcolor{blue}{Kim Darby as Mattie Ross and John Wayne as U.S. Mar-}\\ \textcolor{blue}{shal Rooster Cogburn}. Wayne won his only Academy \\ Award for his performance in this film and reprised his \\ role for the 1975 sequel Rooster Cogburn.\\ \\ \textbf{Document 2}\\ In October 2015, TCM announced the launch of the TCM \\ Wineclub, in which they teamed up with Laithwaite to pro-\\ vide a line of mail-order wines from famous vineyards such \\ as famed writer-director-producer Francis Ford Coppola\textbackslash{}'s \\ winery. Wines are available in 3 month subscriptions, and \\ can be selected as reds, whites, or a mixture of both. From \\ the wines chosen, TCM also includes recommended movies\\  to watch with each, such as a "\textcolor{blue}{True Grit}" wine, to be paired\\  with the \textcolor{blue}{John Wayne} film of the same name.\end{tabular} &
  \begin{tabular}[t]{@{}l@{}}\textbf{Query}\\ In True Grit, who did the star play?\\ \\ \textbf{Document 1}\\ True Grit is a 1969 American western film. It is the first\\  film adaptation of Charles Portis' 1968 novel of the same\\  name. The screenplay was written by Marguerite Roberts.\\  The film was directed by Henry Hathaway and starred \\ Kim Darby as Mattie Ross and John Wayne as U.S. Mar-\\ shal Rooster Cogburn. Wayne won his only Academy \\ Award for his performance in this film and reprised his \\ role for the 1975 sequel Rooster Cogburn.\\ \\ \textbf{Document 2}\\ The Iberian frog, Iberian stream frog or rana patilarga \\ ("Rana iberica") is a species of frog in the family \\ Ranidae found in Portugal and Spain. Its natural ha-\\ bitats are rivers, mountain streams and swamps. It is \\ threatened by habitat loss, introduced species, climate \\ change, water contamination, and increased ultraviolet\\  radiation.\end{tabular} \\ \midrule
\textbf{HotpotQA} &
  \begin{tabular}[t]{@{}l@{}}\textbf{Query}\\ The \textcolor{blue}{2000 British film Snatch} was later adapted into a \\ television series for what streaming service?\\ \\ \textbf{Document 1}\\ \textcolor{blue}{Snatch} is a British/American television series based \\ on the film of the same name, which debuted on \textcolor{blue}{Crackle}\\  on 16 March 2017. The show has been renewed for a \\ second season.\\ \\ \textbf{Document 2}\\ \textcolor{blue}{Snatch} (stylised as snatch.) is a \textcolor{blue}{2000 British crime come-}\\ \textcolor{blue}{dy film} written and directed by Guy Ritchie, featuring an \\ ensemble cast. Set in the London criminal underworld, \\ the film contains two intertwined plots: one dealing with \\ the search for a stolen diamond, the other with a small-\\ time boxing promoter (Jason Statham) who finds himself \\ under the thumb of a ruthless gangster (Alan Ford) who \\ is ready and willing to have his subordinates carry out \\ severe and sadistic acts of violence.\end{tabular} &
  \begin{tabular}[t]{@{}l@{}}\textbf{Query}\\ The 2000 British film Snatch was later adapted into a \\ television series for what streaming service?\\ \\ \textbf{Document 1}\\ Snatch is a British/American television series based \\ on the film of the same name, which debuted on Crackle\\  on 16 March 2017. The show has been renewed for a \\ second season.\\ \\ \textbf{Document 2}\\ Orange Is the New Black (sometimes abbreviated to \\ OITNB) is an American comedy-drama web television\\  series created by Jenji Kohan for Netflix. The series is \\ based on Piper Kerman\textbackslash{}'s memoir, "" (2010), about her\\ experiences at FCI Danbury, a minimum-security federal \\ prison. "Orange Is the New Black" premiered on July 11,\\  2013 on the streaming service Netflix. In February 2016, \\ the series was renewed for a fifth, sixth, and seventh season. \\ The fifth season was released on June 9, 2017. The series \\ is produced by Tilted Productions in association with Lion-\\ sgate Television.\end{tabular} \\ \midrule
\textbf{2Wiki} &
  \begin{tabular}[t]{@{}l@{}}\textbf{Query}\\ Where was the \textcolor{blue}{composer of film Love Story 1999} born?\\ \\ \textbf{Document 1}\\ Devanesan Chokkalingam, popularly known as Deva, is \\ an \textcolor{blue}{Indian film composer} and singer. He has composed \\ songs and provided background music for Tamil, Telugu, \\ Malayalam and Kannada films in a career spanning about \\ 20 years. Many know his gaana songs, written mostly \\ using Madras Tamil. He is known as the "Father of Gaana\\  Genre" in the Tamil film industry. Deva has composed \\ music for many films. He debuted as a film music director\\  in the film "Manasukkeththa Maharaasa" in 1989. In the\\  intervening years he has composed music for a total of \\ more than 400 films.\\ \\ \textbf{Document 2}\\ \textcolor{blue}{Love Story 1999} is a 1998 \textcolor{blue}{Indian} Telugu-language romantic\\  comedy film directed by K. Raghavendra Rao. The film had \\ an ensemble cast starring Prabhu Deva, Vadde Naveen, \\ Ramya Krishna, Rambha and Laila in the lead roles.\end{tabular} &
  \begin{tabular}[t]{@{}l@{}}\textbf{Query}\\ Where was the composer of film Love Story 1999 born?\\ \\ \textbf{Document 1}\\ Devanesan Chokkalingam, popularly known as Deva, is \\ an Indian film composer and singer. He has composed \\ songs and provided background music for Tamil, Telugu, \\ Malayalam and Kannada films in a career spanning about \\ 20 years. Many know his gaana songs, written mostly \\ using Madras Tamil. He is known as the "Father of Gaana\\  Genre" in the Tamil film industry. Deva has composed \\ music for many films. He debuted as a film music director\\  in the film "Manasukkeththa Maharaasa" in 1989. In the\\  intervening years he has composed music for a total of \\ more than 400 films.\\ \\ \textbf{Document 2}\\ Tarcisio Fusco was an Italian composer of film scores. He\\  was the brother of the composer Giovanni Fusco and the \\ uncle of operatic soprano Cecilia Fusco.\end{tabular} \\ \bottomrule
\end{tabular}
\end{table*}

\begin{table*}[t]    
\caption{Cases that the query can be answered correctly in DR-RAG, and can not in Adaptive-RAG. We present wrong answer in \textcolor{red}{red} and the right answer in \textcolor{blue}{blue}.}  
\centering
 \fontsize{8.2}{8.2}\selectfont
\label{some examples_compared_with_adaptive}    
\renewcommand{\arraystretch}{1}%调行距    
\setlength{\tabcolsep}{1.8pt}    
\linespread{1}    
\begin{tabular}{@{}llll@{}}
\toprule
\textbf{Dataset} & \textbf{\textbf{Query} and Answer} & \textbf{Answer by Adaptive-RAG} & \textbf{Answer by DR-RAG} \\ \midrule
\textbf{MuSiQue} & \begin{tabular}[t]{@{}l@{}}\textbf{Query}\\ What is the birth date of the By \\ Your Side performer?\\ \\ \textbf{True Answer}\\ \textcolor{blue}{1 January 1986}.\end{tabular} & \begin{tabular}[t]{@{}l@{}}The performer of "By Your Side" is James\\  Cottriall. James Cottriall was born on \textcolor{red}{20} \\ \textcolor{red}{May 2011}. So the answer is: \textcolor{red}{ 20 May 2011}.\end{tabular} & \begin{tabular}[t]{@{}l@{}}It is mentioned that James Cottriall was \\ born on \textcolor{blue}{1 January 1986}. Additionally, in\\  context, it is stated that "By Your Side" \\ is a song by James Cottriall.Putting this \\ information together, we can determine \\ that the birth date of James Cottriall, the \\ performer of the song "By Your Side," is\\ \textcolor{blue}{1 January 1986}. \\ Answer: <\textcolor{blue}{1 January 1986}>\end{tabular} \\ \midrule
\textbf{HotpotQA} & \begin{tabular}[t]{@{}l@{}}\textbf{Query}\\ Stephen susco wrote a movie \\ produced by which american \\ film actor?\\ \\\textbf{True Answer}\\ \textcolor{blue}{Do Kyung-soo}\end{tabular} & \begin{tabular}[t]{@{}l@{}}The member of the South Korean-Chinese \\ boy group EXO who stars in the upcoming \\ movie "The Underdog" is \textcolor{red}{Sehun}. So the \\ answer is: \textcolor{red}{Sehun}.\end{tabular} & \begin{tabular}[t]{@{}l@{}}It is mentioned that \textcolor{blue}{Do Kyung-soo}, also \\ known by his stage name D.O., is a mem-\\ ber of EXO and has starred in various \\ television dramas and movies. In context,\\  it is mentioned that the upcoming South \\ Korean animated film "The Underdog" stars\\  \textcolor{blue}{Do Kyung-soo}. Therefore, based on the \\ information provided, \textcolor{blue}{Do Kyung-soo} stars\\  in the upcoming movie "The Underdog".\\ Answer: <\textcolor{blue}{Do Kyung-soo}>\end{tabular} \\ \midrule
\textbf{2Wiki} & \begin{tabular}[t]{@{}l@{}}\textbf{Query}\\ Where was the place of death\\  of the director of film The First\\  Day Of Freedom?\\ \\ \textbf{True Answer}\\ \textcolor{blue}{Florida}\end{tabular} & \begin{tabular}[t]{@{}l@{}}The director of The First Day of Freedom is\\ Aleksander Ford. Aleksander Ford died in \\\textcolor{red}{ Warsaw, Poland}. So the answer is: \textcolor{red}{Warsaw,} \\ \textcolor{red}{Poland}.\end{tabular} & \begin{tabular}[t]{@{}l@{}}In context, it is mentioned that Aleksander \\ Ford committed suicide in 1980 in Naples, \\ \textcolor{blue}{Florida}, United States. Given the information\\  provided, the place of death of Aleksander \\ Ford, the director of the film "The First Day \\ of Freedom," was Naples, \textcolor{blue}{Florida}, United \\ States. \\ Answer: <Naples, \textcolor{blue}{Florida}, United States.>\end{tabular} \\ \bottomrule
\end{tabular}
\end{table*}

\begin{table*}[t]    
\caption{A case of documents retrieved by QDC and CFS on the MuSiQue dataset, where the necessary documents are in blue, and the top-$k$ is 4.}    
\centering
\footnotesize    
\label{some examples_dqc_cfs}    
\renewcommand{\arraystretch}{1}%调行距    
\setlength{\tabcolsep}{1.8pt}    
\linespread{1}    
\begin{tabular}{@{}lll@{}}
\toprule
\textbf{Query} & \textbf{Documents retrieved by QDC} & \textbf{Documents retrieved by CFS} \\ \midrule
\begin{tabular}[t]{@{}l@{}}Who is the \textcolor{blue}{spouse} of the \\ \textcolor{blue}{Green} \textcolor{blue}{performer}?\end{tabular} & \begin{tabular}[t]{@{}l@{}}\textbf{Document 1:}\\ {\textcolor{blue}{Green} is the fourth studio album by British pro-}\\ {gressive rock \textcolor{blue}{musician} \textcolor{blue}{Steve Hillage}. Written }\\ {in spring 1977 at the same time as his previous }\\ {album, the funk-inflected "Motivation Radio" }\\ {(1977), "Green" was originally going to be re-}\\{leased as "The Green Album" as a companion} \\ {to "The Red Album" (the originally intended }\\{ name for "Motivation Radio"). However, this }\\ {plan was dropped and after a US tour in late }\\{1977, "Green" was recorded alone, primarily} \\{in Dorking, Surrey, and in London.}\\ \\ \textbf{Document 2:}:\\ "Little Green" is a song composed and performed\\ by Joni Mitchell. It is the third track on her 1971 \\ album "Blue".\\ \\ \textbf{Document 3:}:\\ The Main Attraction is an album by American \\ jazz guitarist Grant Green featuring performances\\ recorded in 1976 and released on the Kudu label.\\ \\ \textbf{Document 4:}:\\ Grant's First Stand is the debut album by Ameri-\\ can jazz guitarist Grant Green featuring perfor-\\ mances by Green recorded and released on the \\ Blue Note label in 1961. Earlier recordings made\\  by Green for Blue Note were released as "First \\ Session" in 2001.\end{tabular} & \begin{tabular}[t]{@{}l@{}}{\textbf{Document 1:}:}\\ {\textcolor{blue}{Green} is the fourth studio album by British pro-}\\ {gressive rock \textcolor{blue}{musician Steve Hillage}. Written }\\ {in spring 1977 at the same time as his previous }\\ {album, the funk-inflected "Motivation Radio" }\\ {(1977), "Green" was originally going to be re-}\\{leased as "The Green Album" as a companion} \\ {to "The Red Album" (the originally intended }\\{ name for "Motivation Radio"). However, this }\\ {plan was dropped and after a US tour in late }\\{1977, "Green" was recorded alone, primarily} \\{in Dorking, Surrey, and in London.}\\ \\ \textbf{Document 2:}:\\ "Little Green" is a song composed and performed\\ by Joni Mitchell. It is the third track on her 1971 \\ album "Blue".\\ \\ {\textbf{Document 3:}:}\\ { \textcolor{blue}{Miquette Giraudy} (born 9 February 1953, Nice, }\\{ France) is a keyboard player and vocalist, best }\\{ known for her work in Gong and with \textcolor{blue}{her partner}}\\ {\textcolor{blue}{Steve Hillage}. She and Hillage currently form the}\\{ core of the ambient band System 7. In addition to}\\{ her performances in music, she has also worked }\\{ as an actress, film editor and writer. In each role,} \\{ she has used different stage names.}\end{tabular}\\ \bottomrule
\end{tabular}
\end{table*}

\begin{table*}[t]    
\caption{A case of documents retrieved by QDC and CIS on the HotpotQA dataset, where the necessary documents are in blue, and the top-$k$ is 4.}
\centering
 \fontsize{8.7}{8.7}\selectfont
\label{some examples_classifier}    
\renewcommand{\arraystretch}{1}%调行距    
\setlength{\tabcolsep}{1.8pt}    
\linespread{1}    
\begin{tabular}{@{}lll@{}}
\toprule
\textbf{Query} & \textbf{Documents retrieved by QDC} & \textbf{Documents retrieved by CIS} \\ \midrule
\begin{tabular}[t]{@{}l@{}}Who is the \textcolor{blue}{child} of \textcolor{blue}{Caro-}\\\textcolor{blue}{line LeRoy's spouse}?  \end{tabular} & \begin{tabular}[t]{@{}l@{}}{\textbf{Document 1:}:}\\ {\textcolor{blue}{Caroline LeRoy} Webster (September 28, 1797 }\\{ in New York City – February 26, 1882) was }\\ {\textcolor{blue}{the second wife} of 19th Century statesman }\\ {\textcolor{blue}{Daniel Webster}. Her father was Herman LeRoy,}\\ { who was once head of the commercial house} \\{ of Leroy, Bayard, McKiven Co., a largetrading }\\ {company that operated in different partsof the }\\{ world. Her father was also the first Holland }\\{ Consul to the United States. Caroline's mother }\\{ was Hannah Cornell, daughter of the last Royal }\\{ Attorney General of the State of North Carolina.}\\ { Caroline was a descendant of Thomas Cornell.}\\ \\ \textbf{Document 2:}:\\ Pierre Paul Leroy-Beaulieu (9 December 1843 \\ in Saumur – 9 December 1916 in Paris) was a \\ French economist, brother of Henri Jean Baptiste\\  Anatole Leroy-Beaulieu, born at Saumur, Maine\\ -et-Loire on 9 December 1843, and educated in \\ Paris at the Lycée Bonaparte and the École de \\ Droit. He afterwards studied at Bonn and Berlin,\\ and on his return to Paris began to write for "Le \\ Temps", "Revue nationale" and "Revue contem-\\ poraine".\\ \\ {\textbf{Document 3:}:}\\{ \textcolor{blue}{Daniel Fletcher Webster}, commonly known as }\\{ Fletcher Webster (July 25, 1813 in Portsmouth, }\\{ New Hampshire – August 30, 1862) was the \textcolor{blue}{son}}\\{ of renowned politician \textcolor{blue}{Daniel Webster} and Grace}\\{ Fletcher Webster. He was educated at Harvard }\\{ College. During his father's first term as Secretary}\\{ of State, Fletcher served as Chief Clerk of the}\\{ United States State Department which, at the time,}\\ {was the second most powerful office in the State }\\{ Department. As Chief Clerk, he delivered the news}\\{ of President William Henry Harrison's death to the }\\{ new President, John Tyler.}\\ \\ \textbf{Document 4:}:\\ Leroy, also Leeroy, LeeRoy, Lee Roy, LeRoy\\ or Le Roy, is both a male given name in English - \\ speaking countries and a family name of French \\ origin. Leroy (lørwa) is one of the most common \\ surnames in northern France. As a surname it is \\ sometimes written Le Roy, as a translation of \\ Breton Ar Roue. It is an archaic spelling of le roi, \\ meaning ``the king ''and is the equivalent of the \\ English surname King.\end{tabular} & \begin{tabular}[t]{@{}l@{}}{\textbf{Document 1:}:}\\{ \textcolor{blue}{Daniel Fletcher} Webster, commonly known as }\\{ Fletcher Webster (July 25, 1813 in Portsmouth,}\\{ New Hampshire – August 30, 1862) was the }\\{ \textcolor{blue}{son} of renowned politician \textcolor{blue}{Daniel Webster} and}\\{ Grace Fletcher Webster. He was educated at }\\{ Harvard College. During his father's first term }\\{ as Secretary of State, Fletcher served as Chief }\\{ Clerk of the United States State Department }\\{ which, at the time, was the second most powerful }\\{ office in the State Department. As Chief Clerk, }\\{ he delivered the news of President William Henry}\\{ Harrison's death to the new President, John Tyler.}\\ \\ {\textbf{Document 2:}:}\\{ \textcolor{blue}{Caroline LeRoy} Webster (September 28, 1797 }\\{ in New York City – February 26, 1882) was }\\ {\textcolor{blue}{the second wife} of 19th Century statesman }\\{ \textcolor{blue}{Daniel Webster}. Her father was Herman LeRoy,}\\{  who was once head of the commercial house }\\ {of Leroy, Bayard, McKiven Co., a largetrading }\\{ company that operated in different partsof the }\\{ world. Her father was also the first Holland} \\{ Consul to the United States. Caroline's mother }\\{ was Hannah Cornell, daughter of the last Royal }\\{ Attorney General of the State of North Carolina.}\\{  Caroline was a descendant of Thomas Cornell.}\\ \\ \\ \textbf{Document 3:}:\\ Leroy, also Leeroy, LeeRoy, \\ Lee Roy, LeRoyor Le Roy, is both a male given \\ name in English - speaking countries and a family \\ name of French origin. Leroy (lørwa) is one of \\ the most common surnames in northern France. \\ As a surname it is sometimes written Le Roy, as \\ a translation of Breton Ar Roue. It is an archaic \\ spelling of le roi, meaning ``the king ''and is the \\ equivalent of the English surname King.\end{tabular} \\ \bottomrule
\end{tabular}
\end{table*}

\end{document}